%% file: main.tex
\documentclass[sigconf,arxiv]{acmart}

\AtBeginDocument{%
  }

\usepackage{algorithm}
\usepackage{algorithmic}
\usepackage{amsmath}
\usepackage{amsfonts}
\usepackage{multirow}

\begin{document}

\title{CAGS: Open-Vocabulary 3D Scene Understanding with Context-Aware Gaussian Splatting}

\author{Wei Sun}
\affiliation{%
  \institution{University of Chinese Academy of Sciences}
  \city{Beijing}
  \country{China}}
\email{sunwei162@mails.ucas.ac.cn}

\author{Yanzhao Zhou}
\affiliation{%
  \institution{University of Chinese Academy of Sciences}
  \city{Beijing}
  \country{China}}
\email{zhouyanzhao@ucas.ac.cn}

\author{Jianbin Jiao}
\affiliation{%
  \institution{University of Chinese Academy of Sciences}
  \city{Beijing}
  \country{China}}
\email{jiaojb@ucas.ac.cn}

\author{Yuan Li$^{\dagger}$}
\affiliation{%
  \institution{University of Chinese Academy of Sciences}
  \city{Beijing}
  \country{China}}
\email{liyuan23@ucas.ac.cn}
\thanks{$^\dagger$ Corresponding author}

\renewcommand{\shortauthors}{Wei Sun et al.}

\begin{abstract}
Open-vocabulary 3D scene understanding is crucial for applications requiring natural language-driven spatial interpretation, such as robotics and augmented reality. While 3D Gaussian Splatting (3DGS) offers a powerful representation for scene reconstruction, integrating it with open-vocabulary frameworks reveals a key challenge: \textit{cross-view granularity inconsistency}. This issue, stemming from 2D segmentation methods like SAM, results in inconsistent object segmentations across views (e.g., a "coffee set" segmented as a single entity in one view but as "cup + coffee + spoon" in another). Existing 3DGS-based methods often rely on isolated per-Gaussian feature learning, neglecting the spatial context needed for cohesive object reasoning, leading to fragmented representations. We propose \textit{Context-Aware Gaussian Splatting (CAGS)}, a novel framework that addresses this challenge by incorporating spatial context into 3DGS. CAGS constructs local graphs to propagate contextual features across Gaussians, reducing noise from inconsistent granularity, employs mask-centric contrastive learning to smooth SAM-derived features across views, and leverages a precomputation strategy to reduce computational cost by precomputing neighborhood relationships, enabling efficient training in large-scale scenes. By integrating spatial context, CAGS significantly improves 3D instance segmentation and reduces fragmentation errors on datasets like LERF-OVS and ScanNet, enabling robust language-guided 3D scene understanding.
\end{abstract}


\maketitle

\input{sec/1_introduction}
\input{sec/2_related_work}
\input{sec/3_preliminaries}

\input{sec/4_method}

\input{sec/5_experiment}
\input{sec/6_conclusion}

\bibliographystyle{ACM-Reference-Format}
\bibliography{main}

\end{document}

%% file: sec/1_introduction.tex
\section*{Introduction}
\begin{figure}[h]
    \centering
    \includegraphics[width=0.45\textwidth]{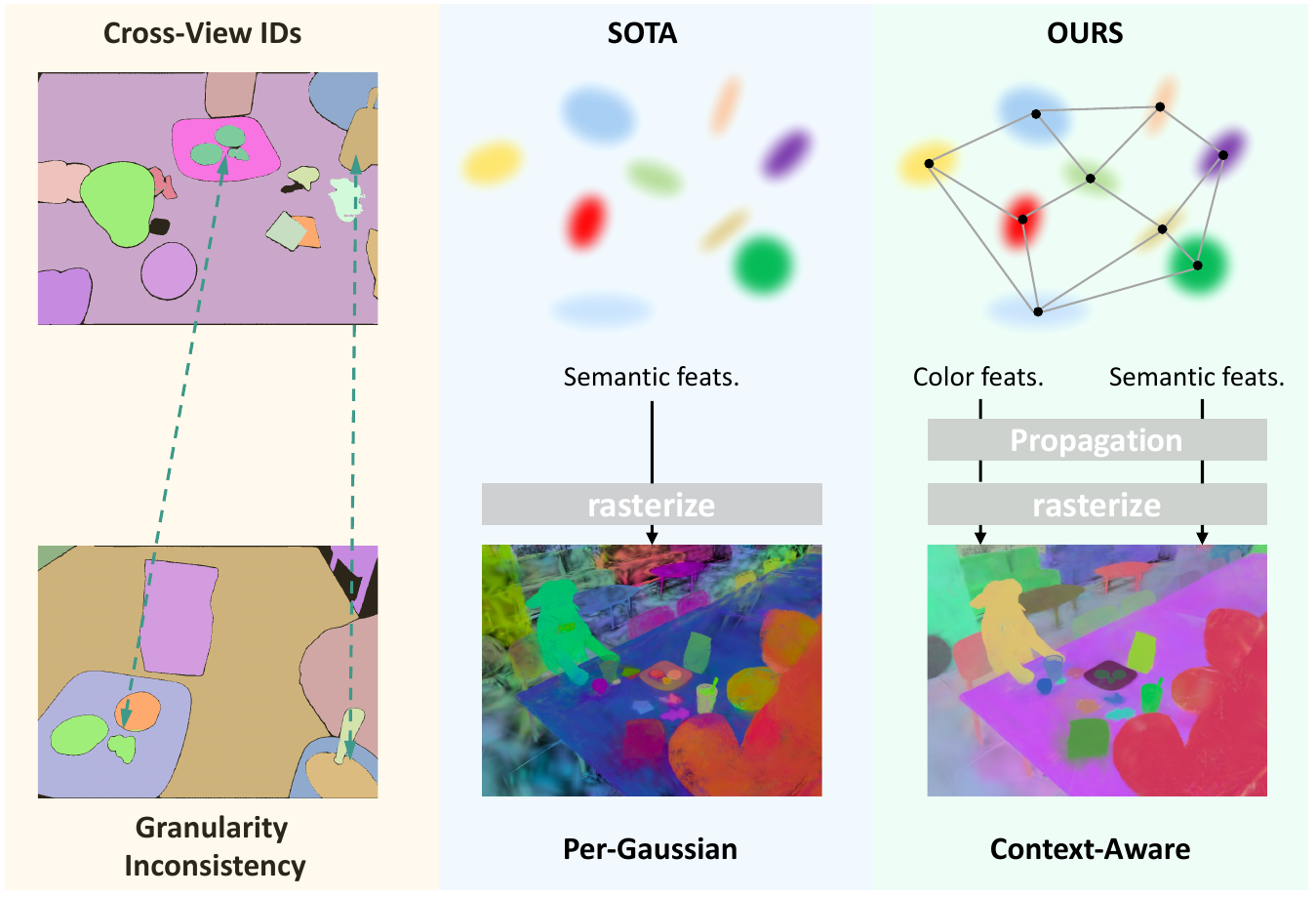}
    \caption{\textnormal{Illustration of cross-view granularity inconsistency in 3DGS-based open-vocabulary scene understanding. A "plate of cookies" may be segmented as a single entity in one view but split into "cookie + cookie + cookie" in another, and a "coffee set" may appear unified in one view but fragmented into "cup + coffee + spoon" in another.}}
    \label{fig:teaser}
\end{figure}

Open-vocabulary 3D scene understanding enables systems to interpret spatial environments through natural language, transcending predefined categories—a capability essential for applications like autonomous navigation, robotics, and augmented reality. While traditional methods often rely on point clouds, the recent rise of 3D Gaussian Splatting (3DGS)~\cite{kerbl20233d} offers a continuous, explicit representation that excels in 3D reconstruction~\cite{keetha2024splatam,lee2024compact,shi2023gir}, 4D reconstruction~\cite{li2024spacetime,luiten2024dynamic,wu20244d,yang2024deformable}, rendering~\cite{sun2024uncertainty,sun2024correspondence} and generation~\cite{tang2023dreamgaussian,ling2024align,yi2024gaussiandreamer}. However, integrating 3DGS with open-vocabulary frameworks exposes a critical challenge: \textit{cross-view granularity inconsistency}. This issue arises from 2D segmentation methods like SAM~\cite{kirillov2023segment}, which lack 3D spatial awareness and produce inconsistent object segmentations across viewpoints. For instance, as illustrated in Fig.~\ref{fig:teaser}, a "plate of cookies" might be segmented as a single entity in one view but split into individual "cookie + cookie + cookie" in another, or a "coffee set" might appear unified in one perspective but fragmented into "cup + coffee + spoon" in another.

Existing 3DGS-based approaches mitigate instance ID inconsistencies—where the same object is assigned different IDs across views—through techniques like cross-view tracking~\cite{ye2024gaussian} or contrastive learning~\cite{wu2024opengaussian,li2024instancegaussian,jun2025dr}. However, they fail to address cross-view granularity inconsistency, as their isolated per-Gaussian feature learning neglects the spatial context necessary for cohesive object reasoning. This results in fragmented representations that hinder comprehensive scene understanding.

To overcome this, we introduce \textit{Context-Aware Gaussian Splatting (CAGS)}, a novel framework that enhances 3DGS by incorporating spatial context to tackle cross-view granularity inconsistency. Through \textit{Contextual Feature Propagation}, CAGS constructs localized graphs on Gaussian primitives, enabling lightweight feature propagation. This allows each Gaussian to aggregate spatial information from its neighbors, reducing noise from inconsistent granularity and enhancing feature smoothness within visually unified objects. Compared with per-pixel semantic learning, our method introduces minimal extra training overhead, leading to faster convergence and improved feature quality.

To further address granularity mismatches, CAGS employs \textit{Mask-Aware Contrastive Learning}. Unlike prior methods that rely on per-pixel InfoNCE losses~\cite{he2020momentum}—vulnerable to granularity noise—our approach computes feature centroids within SAM-derived masks and applies contrastive supervision to these centroids across views. This reduces the impact of inconsistent segmentation, stabilizes training, and preserves feature consistency within objects.

For efficient training in large-scale scenes, CAGS introduces a \textit{Precomputation for Efficient Training} strategy. By precomputing neighborhood relationships after freezing Gaussian positions, we eliminate redundant graph computations during training, significantly reducing computational cost while maintaining accuracy. This enables CAGS to scale effectively to scenes with millions of Gaussians, such as LERF-OVS~\cite{ye2024gaussian}.

Finally, through \textit{Instance Clustering and Semantic Matching}, CAGS associates 3D instances with 2D open-vocabulary priors, ensuring robust semantic alignment across views. Our experiments on LERF-OVS and ScanNet~\cite{dai2017scannet} demonstrate CAGS’s effectiveness, with significant improvements in 3D instance segmentation and reduced fragmentation errors. These results highlight the critical role of spatial context in enabling accurate open-vocabulary 3D understanding.

\textbf{Contributions:}
\begin{itemize}
    \item A \textit{Contextual Feature Propagation} framework that mitigates cross-view granularity inconsistency in 3DGS.
    \item A \textit{Mask-Aware Contrastive Learning} strategy that supervises mask centroids, reducing noise from inconsistent granularity.
    \item A \textit{Precomputation for Efficient Training} strategy that reduces computational cost by precomputing neighborhood relationships.
    \item Significant improvements in 3D instance segmentation and reduced fragmentation errors on LERF-OVS and ScanNet.
\end{itemize}

%% file: sec/2_related_work.tex
\section{Related Work}
\subsection{Neural Rendering Techniques}  
The evolution of neural scene representations has been significantly shaped by Neural Radiance Fields (NeRF)~\cite{mildenhall2021nerf}, which pioneered photorealistic novel view synthesis through differentiable volume rendering. Subsequent improvements in rendering fidelity have further advanced the field by introducing hybrid architectural designs~\cite{barron2021mip,barron2022mip,barron2023zip}. However, the inherent computational complexity of volumetric integration in NeRF-based approaches imposes constraints on training efficiency and real-time applicability. This challenge has driven the adoption of explicit scene representations, including voxel-based structures~\cite{sun2022direct,fridovich2022plenoxels,reiser2023merf}, hash-encoded feature grids~\cite{muller2022instant}, and point cloud systems~\cite{yifan2019differentiable,aliev2020neural,ruckert2022adop}, which prioritize computational efficiency through spatial discretization. The emergence of 3D Gaussian Splatting (3DGS)~\cite{kerbl20233d} marks a paradigm shift by replacing traditional ray marching with differentiable rasterization of anisotropic 3D Gaussians, achieving unprecedented rendering speeds while preserving visual quality. Building upon these foundations, our work extends 3DGS's capabilities toward fine-grained semantic understanding at the point level.

\subsection{Open-Vocabulary 3D Understanding}  
The integration of vision-language models (VLMs) with 3D scene analysis has catalyzed progress in open-vocabulary understanding. Early methodologies~\cite{huang2023clip2point,xue2024ulip,zhu2023pointclip} focused on aligning 3D geometric data with 2D semantic projections from models like CLIP~\cite{radford2021learning} and DINO~\cite{caron2021emerging}, though often struggling with cross-view consistency. The advent of the Segment Anything Model (SAM)~\cite{kirillov2023segment} accelerated the fusion of semantic segmentation with neural rendering, exemplified by LERF~\cite{kerr2023lerf} and subsequent NeRF-based approaches~\cite{liu2023weakly} that distill 2D VLM features into volumetric representations. Recent efforts have transitioned toward 3DGS-centric frameworks~\cite{cen2023segment,choi2024click} to overcome NeRF's computational limitations: LEGaussians~\cite{shi2024language} augments Gaussians with uncertainty-aware semantic attributes for mask rendering, while LangSplat~\cite{qin2024langsplat} employs scene-specific autoencoders to enhance object boundary delineation. Feature3DGS~\cite{zhou2024feature} introduces multi-head rasterizers for joint geometry-feature reconstruction, and Gaussian Grouping~\cite{ye2024gaussian} introduces a tracking model to alleviate the cross-view instance id inconsistency. OpenGaussian~\cite{wu2024opengaussian} employs a two-stage coarse-to-fine codebook to discretize instance features for point-level open-vocabulary 3D understanding. Dr. Splat~\cite{jun2025dr} uses product quantization to compactly represent language-aligned embeddings in 3D Gaussians, bypassing rendering for efficient 3D perception, while one parallel innovation~\cite{zhu2025rethinking} introduces a global object-level codebook for end-to-end 3D scene segmentation, directly associating object-level features with Gaussian points during training. While these methods have significantly promoted open-vocabulary 3D understanding, and among them, some have gone a step further to mitigate the cross-view instance id inconsistency by introducing tracking~\cite{ye2024gaussian} and contrastive learning techniques~\cite{wu2024opengaussian,jun2025dr,zhu2025rethinking,li2024instancegaussian}, they still struggle with cross-view granularity inconsistency due to their lack of spatial context. This deficiency leads to fragmented 3D object representations. In contrast, our CAGS framework takes advantage of context-aware propagation to guarantee more accurate language-grounded 3D scene understanding.

%% file: sec/3_preliminaries.tex
\section{Preliminaries}

\subsection{3D Gaussian Splatting Fundamentals}
3D Gaussian Splatting (3DGS)~\cite{kerbl20233d} represents 3D scenes through a collection of anisotropic Gaussian primitives, each defined by a positional mean $\mu = [x_\mu, y_\mu, z_\mu]^\top$ and a covariance matrix $\Sigma_{3D} \in \mathbb{R}^{3\times3}$ that controls the spatial extent of the Gaussian. The covariance matrix is decomposed into a rotation matrix $R \in SO(3)$ and a diagonal scale matrix $S = \text{diag}(s_x, s_y, s_z)$ through the factorization $\Sigma_{3D} = RSS^\top R^\top$. Each Gaussian additionally carries visual attributes including opacity $\alpha \in [0,1]$ and RGB color $c \in \mathbb{R}^3$, collectively parameterizing the scene as $\Theta = \{\mu_i, S_i, R_i, \alpha_i, c_i\}_{i=1}^N$. 

The rendering process synthesizes 2D pixel colors $\hat{c}$ through alpha compositing:
\begin{equation}
    \hat{c} = \sum_{i=1}^N T_i \tilde{\alpha}_i c_i, \quad \tilde{\alpha}_i = \alpha_i \exp\left(-\frac{1}{2}d_i^\top \Sigma_{2D}^{-1}d_i\right),
\end{equation}
where $T_i$ denotes transmittance, $d_i \in \mathbb{R}^2$ measures pixel-space distance to Gaussian centers, and $\Sigma_{2D}$ represents projected 2D covariance. The parameters \(\Theta\) of the 3D Gaussian scene are adjusted to minimize the rendering loss between the input image color \(\mathbf{c}\) and the rendered color \(\hat{\mathbf{c}}(\theta)\), as defined in Eq. (1) by the expression \(\arg \min_{\theta} \|\mathbf{c} - \hat{\mathbf{c}}(\theta)\|^2_F\).

\begin{figure*}
    \centering
    \includegraphics[width=\textwidth]{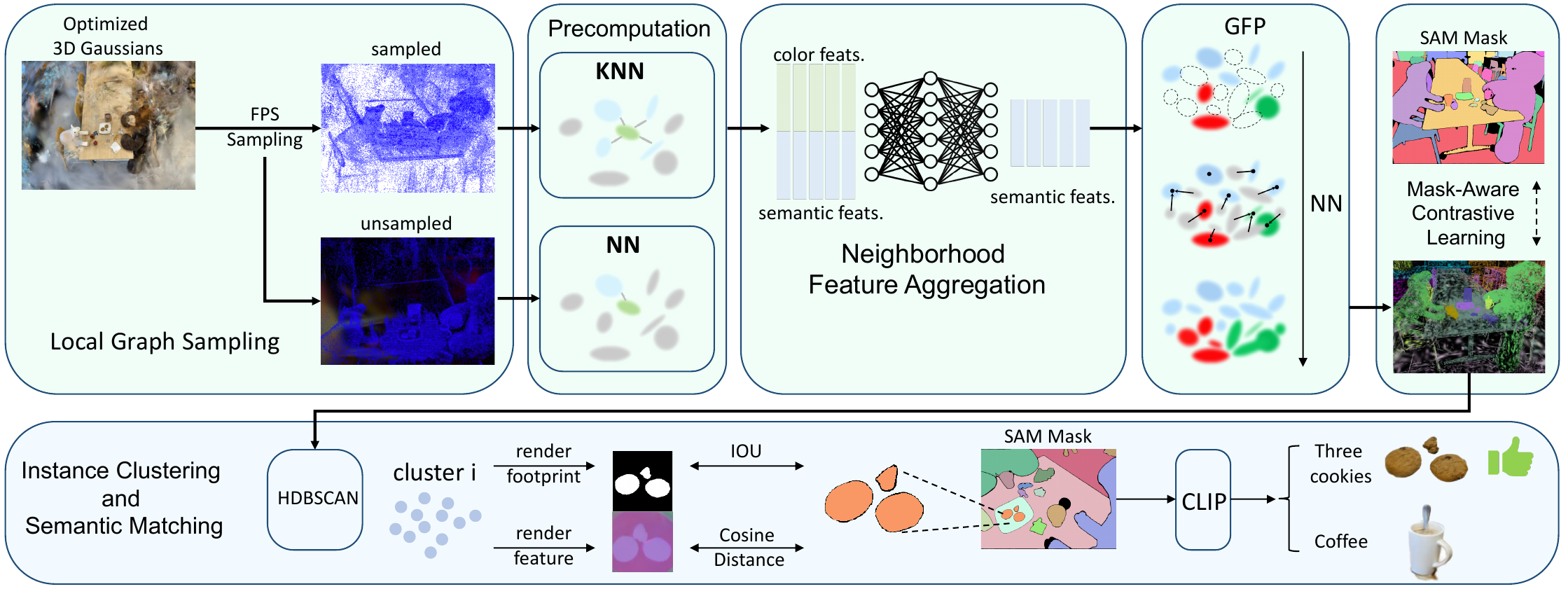}
    \caption{\textnormal{Overview of the Context-Aware Gaussian Splatting (CAGS) pipeline for open-vocabulary 3D scene understanding, starting with optimized 3D Gaussians. The pipeline includes: (1) Local Graph Sampling, (2) Neighborhood Feature Aggregation, (3) Global Feature Propagation (GFP), (4) Mask-Aware Contrastive Learning, and (5) Instance Clustering and Semantic Matching.}}
    \label{fig:pipeline}
\end{figure*}

\subsection{Semantic Feature Embedding in 3DGS}
Recent extensions augment 3DGS with language understanding by replacing color rendering with semantic feature projection~\cite{qin2024langsplat,shi2024language}. Each Gaussian is augmented with language embeddings $\tilde{f}_i \in \mathbb{R}^d$, yielding an extended parameterization $\Phi = \{\mu_i, S_i, R_i, \alpha_i, c_i, \tilde{f}_i\}_{i=1}^N$. Analogous to color synthesis, language features are rendered through:
\begin{equation}
    \hat{f} = \sum_{i=1}^N T_i \tilde{\alpha}_i \tilde{f}_i,
\end{equation}
where $\hat{f}$ denotes the projected 2D semantic feature map. With the aid of feature rendering techniques, recent methods~\cite{ye2024gaussian,wu2024opengaussian,jun2025dr,zhu2025rethinking} employ CLIP~\cite{radford2021learning} to realize open-vocabulary scene understanding.

%% file: sec/4_method.tex
\section{Methodology}

\subsection{Contextual Feature Propagation}

Traditional per-Gaussian feature learning in 3D Gaussian Splatting (3DGS)~\cite{ye2024gaussian,wu2024opengaussian,jun2025dr,zhu2025rethinking} often fails to incorporate the spatial context required for open-vocabulary 3D understanding, leading to fragmented feature representations under cross-view granularity inconsistency. While 3DGS excels in photorealistic rendering through precise control of color and position attributes of individual Gaussians, effective scene interpretation demands aggregating spatial context across related regions to ensure feature smoothness within visually unified objects. A key insight lies in the distinction between photorealistic and semantic rendering: photorealistic rendering captures rich, varied textures even within a single object, making color attributes well-suited for per-Gaussian learning; in contrast, semantic space is relatively sparse, as an object’s internal semantics tend to be consistent, necessitating stronger exploitation of spatial adjacency among Gaussians. However, assigning uniform semantics via codebooks~\cite{wu2024opengaussian,zhu2025rethinking} may oversimplify, as cross-view granularity inconsistency requires feature flexibility to allow subtle variations within an object. Thus, we design a context-aware feature propagation mechanism using local graph construction to model semantic sparsity while preserving feature elasticity for robust 3D understanding, ensuring that the process integrates both spatial relationships and semantic coherence effectively. The method overview is shown in Fig.~\ref{fig:pipeline}.

\noindent\textbf{Local Graph Sampling.} The process starts by constructing a sparse graph to efficiently capture the scene’s topological structure, which is essential for enabling context-aware feature propagation in 3DGS. We employ farthest point sampling (FPS)~\cite{qi2017pointnet} to select 10\% of the total Gaussians ($M = \lfloor 0.1N \rfloor$) as anchors, forming a skeletal structure that preserves the spatial distribution of the scene, where $N$ denotes the total number of Gaussians and $M$ the number of anchors. This sampling strategy ensures a balanced representation by prioritizing distant points, which is particularly important in scenes with uneven Gaussian distributions, such as those containing both densely packed objects (e.g., a cluttered kitchen counter with multiple utensils) and sparse background areas (e.g., an open floor space or a distant wall). By maintaining the overall geometric integrity of the scene, FPS prevents over-sampling in crowded regions, ensuring that the graph captures a representative subset of the scene’s structure across diverse spatial configurations. Each anchor $a_m$ is then connected to its $k=16$ nearest neighbors based solely on Euclidean positions $\mu$:
\begin{equation}
d_{mn} = \|\mu_m - \mu_n\|_2
\end{equation}
where $d_{mn}$ represents the distance between anchors $a_m$ and $a_n$, and $\mu_m$ and $\mu_n$ are the 3D positions of Gaussians $m$ and $n$, respectively. This spatial adjacency ensures that connections reflect geometric proximity, which is critical for modeling the consistent semantics within an object, as spatially close Gaussians are more likely to belong to the same semantic entity. The choice of $k=16$ is empirically determined to balance the trade-off between capturing sufficient local context and minimizing computational overhead, ensuring that the graph remains sparse yet informative for subsequent feature aggregation, especially in large-scale scenes with millions of Gaussians.

\noindent\textbf{Neighborhood Feature Aggregation.} Within the anchor neighborhood, we aggregate features by incorporating RGB colors $c$ alongside semantic features to enhance contextual awareness, a key step in addressing cross-view granularity inconsistency. We first extract the original semantic features $f$ and color features $c$, detaching them from the computational graph to prevent gradient updates to the original features:
\begin{equation}
f^{\text{init}} = \text{Detach}(f), \quad c^{\text{init}} = \text{Detach}(c), \quad h_n = [f^{\text{init}}_n \| c^{\text{init}}_n]
\end{equation}
where $f^{\text{init}}_n$ and $c^{\text{init}}_n$ are the detached features for Gaussian $n$, and $h_n$ is the concatenated feature vector. The concatenated features $h_n$ are then normalized to ensure stable training:
\begin{equation}
h_n^{\text{norm}} = \frac{h_n - \text{mean}(h_n)}{\text{std}(h_n) + \epsilon}, \quad \epsilon = 10^{-10}
\end{equation}
where $h_n^{\text{norm}}$ is the normalized concatenated feature, and $\epsilon$ is a small constant for numerical stability. These normalized features are fed into a GNN~\cite{qi2017pointnet++} for aggregation over two layers:
\begin{equation}
f_m^{(l+1)} = \text{ReLU}\left(\mathbf{W}_1 f_m^{(l)} + \sum_{n\in\mathcal{N}(m)} \mathbf{W}_2 h_n^{\text{norm}}\right)
\end{equation}
where $f_m^{(l)}$ and $f_m^{(l+1)}$ denote the features of anchor $m$ at layer $l$ and $l+1$, respectively, $\mathcal{N}(m)$ is the set of neighboring Gaussians of anchor $m$, and $\mathbf{W}_1$ and $\mathbf{W}_2$ are learnable weight matrices. Notably, the GNN used in our feature propagation employs an intermediate channel size of 16, balancing computational efficiency and feature expressiveness. This inclusion of color information strengthens aggregation by providing additional visual cues, as color attributes often correlate with semantic boundaries (e.g., a white cup and brown coffee within a coffee set), aiding in disambiguating objects with similar semantics but distinct appearances. The two-layer aggregation captures both immediate and extended local contexts, effectively reducing noise from inconsistent segmentation across views caused by cross-view granularity variations, by iteratively refining semantic features within spatially adjacent Gaussians through a GNN, where learnable weights $\mathbf{W}_1$ and $\mathbf{W}_2$ adaptively balance self-features and neighbor contributions, while the concatenated color features provide visual cues to preserve semantic boundaries at object interfaces, ensuring that smoothing occurs primarily within objects and mitigates view-dependent segmentation discrepancies without blending features across distinct objects. The two-layer aggregation results in enriched features that better reflect the local spatial context, enhancing the overall robustness of the semantic representation.

\noindent\textbf{Global Feature Propagation.} Finally, we propagate the aggregated features from anchors to all Gaussians using nearest-neighbor interpolation to ensure a consistent enhancement of semantic features across the entire scene, addressing the disparity where only sampled anchors benefit from GNN-based aggregation while unsampled Gaussians retain their initial per-Gaussian features without contextual refinement. This propagation assigns the enriched features from anchors to all Gaussians based on spatial proximity, ensuring that all points within a spatially coherent region share a uniformly enhanced semantic representation, which mitigates the impact of cross-view granularity variations by maintaining semantic consistency across views.
\begin{equation}
\hat{f}_i = f_{m^*}^{(2)}, \quad m^* = \arg\min_{m \in \mathcal{A}} \|\mu_i - \mu_m\|_2
\end{equation}
where $\hat{f}_i$ is the propagated feature for Gaussian $i$, $f_{m^*}^{(2)}$ is the feature of the nearest anchor $m^*$ after two-layer aggregation, $\mathcal{A}$ is the set of anchors, and $\mu_i$ and $\mu_m$ are the positions of Gaussian $i$ and anchor $m$, respectively. The propagated features are then normalized:
\begin{equation}
\hat{f}_i^{\text{norm}} = \frac{\hat{f}_i - \text{mean}(\hat{f}_i)}{\text{std}(\hat{f}_i) + \epsilon}, \quad \epsilon = 10^{-10}
\end{equation}
where $\hat{f}_i^{\text{norm}}$ is the normalized propagated feature. A residual connection combines the normalized propagated features with the original features:
\begin{equation}
f_i^{\text{final}} = f_i + \hat{f}_i^{\text{norm}}
\end{equation}
where $f_i$ is the original feature of Gaussian $i$, and $f_i^{\text{final}}$ is the final enhanced feature. This design ensures training stability across scenes with Gaussian densities ranging from 10K to over 1M primitives, such as the compact ScanNet dataset and the denser LERF-OVS dataset, by preserving the original learned semantics while allowing contextual enhancements. By integrating spatial context in this manner, our propagation mechanism enhances the robustness of 3DGS for open-vocabulary understanding, enabling more accurate and cohesive scene interpretation in diverse scenarios.

\subsection{Mask-Aware Contrastive Learning}  
The inherent granularity inconsistency of SAM-generated masks across different viewpoints poses a significant challenge for open-vocabulary 3D understanding. Traditional pixel-level contrastive learning approaches~\cite{wu2024opengaussian,jun2025dr,zhu2025rethinking}, which enforce similarity between individual pixels, amplify segmentation noise caused by SAM's fragmentation—such as partial occlusions or varying object part segmentations across views. To address this, we propose a mask-level contrastive learning strategy that leverages the intrinsic coherence of object instances, even when their 2D masks are inconsistently fragmented, by applying a per-mask InfoNCE loss, in contrast to the per-pixel InfoNCE loss~\cite{zhu2025rethinking} used in prior methods, to better capture instance-level semantics.

For each SAM mask \( M_k \) in a training view, we compute a feature centroid by averaging the rendered features \( \hat{f}(u,v) \) across all pixels \( (u,v) \) within the mask:
\begin{equation}
\bar{f}_k = \frac{1}{|M_k|} \sum_{(u,v)\in M_k} \text{Render}(\hat{f}, u,v)
\end{equation}
where \( \bar{f}_k \) is the feature centroid for mask \( M_k \), \( |M_k| \) is the number of pixels in the mask, and \( \text{Render}(\hat{f}, u,v) \) denotes the rendered feature at pixel \( (u,v) \) derived from the Gaussian features \( \hat{f} \). This centroid aggregation acts as a low-pass filter, suppressing pixel-level segmentation errors while preserving the dominant semantic signal of the instance. We then apply a per-mask InfoNCE loss on the centroids \( \bar{f}_k \) of all masks within the same view:
\begin{equation}
\mathcal{L}_{\text{cont}} = -\log \frac{\exp(\text{sim}(\bar{f}_k, \bar{f}_k^+) / \tau)}{\sum_{j=1, j \neq k}^K \exp(\text{sim}(\bar{f}_k, \bar{f}_j) / \tau)}
\end{equation}
where \( \mathcal{L}_{\text{cont}} \) is the contrastive loss, \( \bar{f}_k \) uses its own centroid as the positive pair (self-similarity without augmentation), \( \bar{f}_j \) represents the centroid of other masks \( j \) (negative samples) in the same view, \( K \) is the total number of masks in the view, \( \text{sim}(\cdot, \cdot) \) computes the cosine similarity, and \( \tau \) is a temperature parameter controlling the sharpness of the similarity distribution. By operating at the mask level within a single view, this approach leverages the smoothing effect of centroid aggregation to reduce noise inherent in per-pixel contrastive losses, making the learned features more robust to SAM’s inconsistent fragmentation, such as varying granularity in object part segmentations. This also aligns with the sparse nature of semantic space, as the per-mask loss encourages distinct feature separation between different instances while maintaining intra-instance consistency, enhancing the model’s ability to handle segmentation discrepancies.

\subsection{Precomputation for Efficient Training}  
The computational efficiency of our framework stems from a critical design choice: after the initial photometric optimization phase converges, we freeze the Gaussian positions $\mu_i$ and disable densification/pruning operations. This spatial invariance enables one-time precomputation of all neighborhood relationships prior to semantic feature learning, eliminating redundant graph computations during training iterations.  

The precomputation involves two key steps. First, for the sampled anchors (selected as 10\% of the Gaussians), we compute k-nearest neighbor (k=16) graphs using Euclidean distances, accelerated through spatial hashing techniques, to establish the neighborhood relationships for subsequent GNN-based aggregation. Second, for each unsampled Gaussian, we identify its nearest anchor among the sampled set, using an efficient search within the subsampled anchor set rather than a global search across all Gaussians. These relationships are then stored for use throughout training.

This precomputation strategy yields two key advantages. First, by storing both the k-NN graphs for the sampled anchors ($O(Mk)$ memory, where $M$ is the number of anchors) and the nearest anchor indices for unsampled Gaussians ($O(N)$ memory, where $N$ is the total number of Gaussians), we efficiently manage memory usage while enabling fast access to neighborhood relationships. Notably, the nearest anchor search for unsampled Gaussians is performed only within the subsampled anchor set to ensure they receive the contextual enhancement from sampled anchors, as described in the global feature propagation step, while also reducing the computational cost compared to a global search across all Gaussians. Second, eliminating per-iteration neighbor searches significantly reduces training time; for a scene with 1M Gaussians, the one-time k-NN graph construction and nearest anchor search can be completed in approximately 5 minutes. Crucially, the frozen Gaussian positions guarantee the precomputed graphs remain valid throughout feature learning—recomputing neighborhoods during training would otherwise incur prohibitive $O(N^2)$ costs. This precomputation enables efficient training and feature propagation, striking an optimal balance between accuracy and efficiency.

\begin{algorithm}[t]
\caption{Context-Aware Feature Learning}
\label{alg:feature_learning}
\begin{algorithmic}[1]
\REQUIRE 
  Initial Gaussians $\Phi = \{\mu_i, S_i, R_i, \alpha_i, c_i\}$, Images $\{\mathcal{I}_v\}$, SAM masks $\{\mathcal{M}_v\}$, CLIP encoder $\mathcal{E}_{\text{CLIP}}$
\ENSURE 
  Enhanced Gaussians $\Phi' = \{\mu_i, S_i, R_i, \alpha_i, c_i, f_i^{\text{final}}\}$

\STATE \textbf{Stage 1: Geometric Initialization}  (30000 iterations)
\STATE $\quad$ Optimize $\Phi$ via $\mathcal{L}_{\text{rgb}}$, perform densification/pruning
\STATE $\quad$ Freeze parameters $\{\mu_i, S_i, R_i, c_i, \alpha_i\}$

\STATE \textbf{Stage 2: Contextual Propagation}  (10000 iterations, 20min)
\STATE $\quad$ Precompute anchor graph $\mathcal{G}$:
\STATE $\qquad$ $\mathcal{A} \gets \text{FPS}(\{\mu_i\})$, $\mathcal{N}(a_m) \gets \text{kNN}(\{\mu_i\})$
\STATE $\qquad$ $m^*_i \gets \arg\min_{m \in \mathcal{A}} \|\mu_i - \mu_m\|_2$
\STATE $\quad$ \textbf{For each iteration:}
\STATE $\qquad$ Extract features: $f_n^{\text{init}} \gets \text{Detach}(f_n)$, $c_n^{\text{init}} \gets \text{Detach}(c_n)$
\STATE $\qquad$ Concatenate features: $h_n \gets [f_n^{\text{init}} \| c_n^{\text{init}}]$
\STATE $\qquad$ Normalize features: $h_n^{\text{norm}} \gets \frac{h_n - \text{mean}(h_n)}{\text{std}(h_n) + \epsilon}$
\STATE $\qquad$ Aggregate via GNN: $f_m^{(l+1)} \gets \text{ReLU}\left(\mathbf{W}_1 f_m^{(l)} + \sum_{n\in\mathcal{N}(m)} \mathbf{W}_2 h_n^{\text{norm}}\right)$
\STATE $\qquad$ Propagate features: $\hat{f}_i \gets f_{m^*_i}^{(2)}$
\STATE $\qquad$ Normalize features: $\hat{f}_i^{\text{norm}} \gets \frac{\hat{f}_i - \text{mean}(\hat{f}_i)}{\text{std}(\hat{f}_i) + \epsilon}$
\STATE $\qquad$ Update features: $f_i^{\text{final}} \gets f_i + \hat{f}_i^{\text{norm}}$
\STATE $\qquad$ Feature rendering: $\hat{f} \gets \sum_{i=1}^N T_i \tilde{\alpha}_i f_i^{\text{final}}$,
\STATE $\qquad$ Compute mask centroids: $\bar{f}_k \gets \frac{1}{|M_k|} \sum_{(u,v)\in M_k} \text{Render}(\hat{f}_i, u,v)$
\STATE $\qquad$ Apply contrastive loss: $\mathcal{L}_{\text{cont}} \gets -\log \frac{\exp(\text{sim}(\bar{f}_k, \bar{f}_k)/\tau)}{\sum_{j \neq k} \exp(\text{sim}(\bar{f}_k, \bar{f}_j)/\tau)}$
\STATE $\quad$ \textbf{End for}

\STATE \textbf{Stage 3: Instance Clustering and Semantic Matching}
\STATE $\quad$ Cluster Gaussians: $\mathcal{C} \gets \text{HDBSCAN}\left(\frac{f_i^{\text{final}} - \mu_f}{\sigma_f}\right)$
\STATE $\quad$ Match clusters to language features via IoU and cosine similarity
\end{algorithmic}
\end{algorithm}

\begin{figure*} 
    \centering
    \includegraphics[width=\textwidth]{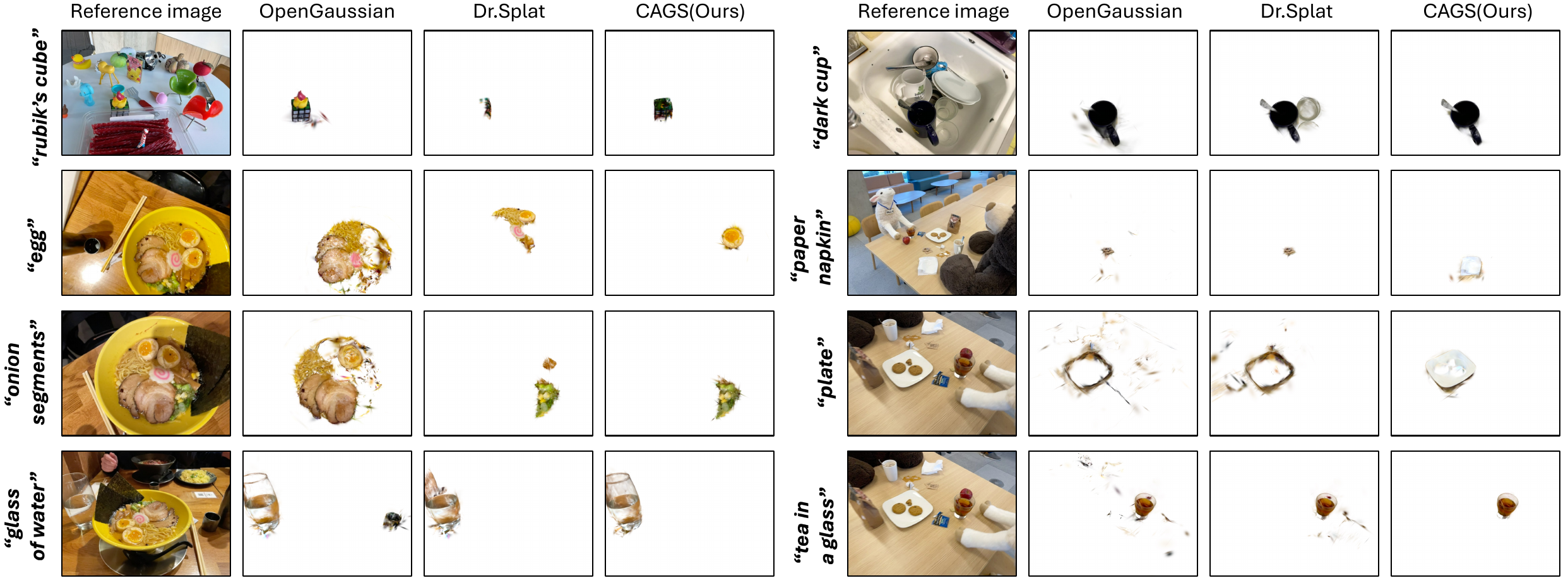}
    \caption{\textnormal{Text query visualization on the LERF-OVS dataset. Columns compare reference images with segmentations from OpenGaussian, Dr.Splat, and our CAGS method for queries like "egg" and "rubik's cube." CAGS achieves more accurate target identification with fewer noisy fragments, avoiding nearby objects and background noise.}}
    \label{fig:lerf}
\end{figure*}

\begin{table*}[htbp]
\centering
\caption{\textnormal{Evaluation of text-driven 3D object selection on the LERF-OVS dataset. Performance is measured using mIoU and mAcc@0.25 across different class settings.}}
\label{tab:lerf}
\resizebox{\textwidth}{!}{
\begin{tabular}{c|ccccc|ccccc}
\toprule
\multirow{2}{*}{Methods} & \multicolumn{5}{c|}{mIoU} & \multicolumn{5}{c}{mAcc @ 0.25} \\
& waldo\_kitchen & ramen & figurines & teatime & Mean & waldo\_kitchen & ramen & figurines & teatime & Mean \\
\midrule
LangSplat~\cite{qin2024langsplat} & 9.18 & 7.92 & 10.16 & 11.38 & 9.66 & 9.09 & 11.27 & 8.93 & 20.34 & 12.41 \\
OpenGaussian~\cite{wu2024opengaussian} & 22.70 & \underline{31.01} & 39.29 & \underline{60.44} & 38.36 & 31.82 & 42.25 & 55.36 & 76.27 & 51.43 \\
Dr.Splat~\cite{jun2025dr} & \underline{37.05} & 24.33 & \underline{54.42} & 57.35 & \underline{43.29} & \textbf{63.64} & \underline{35.21} & \underline{80.36} & \underline{77.97} & \underline{64.30} \\
Ours & \textbf{37.62} & \textbf{36.29} & \textbf{60.85} & \textbf{68.40} & \textbf{50.79} & \textbf{63.64} & \textbf{46.48} & \textbf{82.14} & \textbf{86.44} & \textbf{69.62} \\
\bottomrule
\end{tabular}
}
\end{table*}

\subsection{Instance Clustering and Semantic Matching}  
To associate 3D Gaussians with open-vocabulary language features, we employ a two-stage strategy that integrates feature-based clustering and language feature matching. After context-aware feature learning, we first cluster Gaussians into coherent 3D instances using HDBSCAN~\cite{mcinnes2017hdbscan} in the normalized feature space:
\begin{equation}
\mathcal{C} = \{C_k\} = \text{HDBSCAN}\left(\frac{f_i^{\text{final}} - \mu_f}{\sigma_f}\right)
\end{equation}
where $\mathcal{C}$ is the set of clusters, $C_k$ represents the $k$-th 3D instance, $f_i^{\text{final}}$ is the enhanced feature of Gaussian $i$, and $\mu_f$ and $\sigma_f$ denote the mean and standard deviation of the enhanced features across all Gaussians, respectively.

For each 3D instance $C_k$, we render its spatial footprint $M_k \in \{0,1\}^{H\times W}$ and compute its mean semantic features $\tilde{f}_k \in \mathbb{R}^d$ in the current view by averaging the rendered features across the instance’s silhouette~\cite{wu2024opengaussian}. Concurrently, SAM provides 2D masks $\{B_j\}$, and we compute their associated language features $\{f_j\}$ by first rendering the semantic features of all Gaussians into a feature map in the current view, then filling each SAM mask $B_j$ with this rendered feature map and averaging the features within the mask to obtain $f_j$. We compute association scores combining spatial overlap and semantic consistency:
\begin{equation}
S_{kj} = \underbrace{\text{IoU}(M_k, B_j)}_{\text{Spatial}} \cdot \underbrace{(1 - \text{cosine}(\tilde{f}_k, f_j))}_{\text{Semantic}}
\end{equation}
where $S_{kj}$ is the association score between instance $C_k$ and SAM mask $B_j$ and $\text{cosine}(\cdot, \cdot)$ computes the cosine distance. The instance-mask pair $(C_k, B_j)$ with the maximal $S_{kj}$ in each view is associated, and the mask-level CLIP feature of the matched SAM mask is assigned to the instance. Multi-view consensus is achieved by accumulating the CLIP features of the matched SAM masks across all viewpoints where $C_k$ is observed, then averaging them to obtain a single feature vector per instance, ensuring a robust and consistent semantic representation. Our overall algorithm is presented as Algorithm.~\ref{alg:feature_learning}.

%% file: sec/5_experiment.tex
\section{Experiments}

\begin{figure*} 
    \centering
    \includegraphics[width=\textwidth]{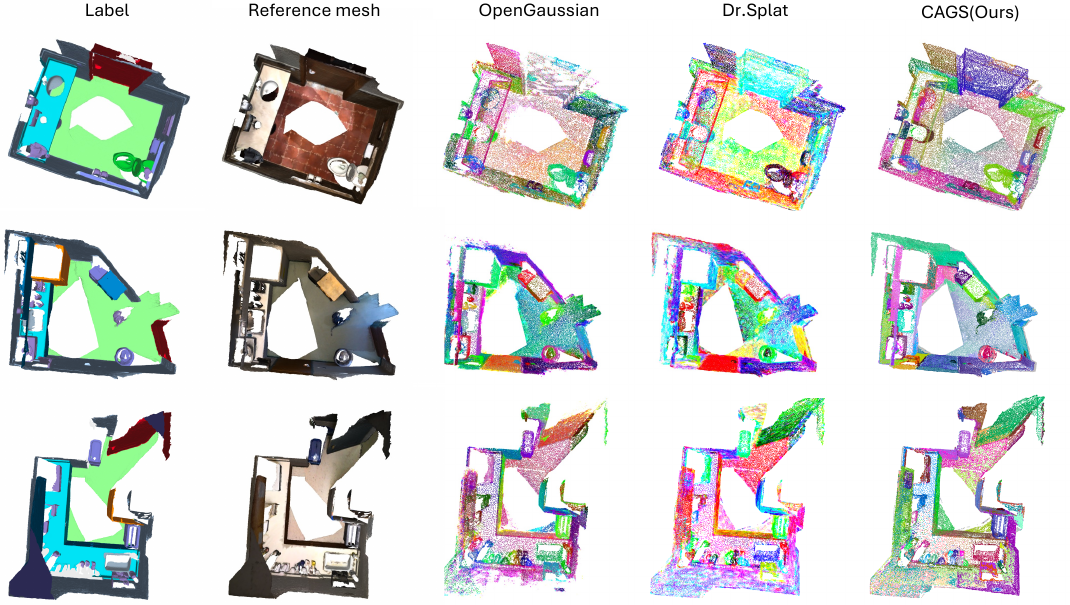}
    \caption{\textnormal{Comparison of feature visualizations on the ScanNet dataset. Rows represent different scenes, highlighting the ability of each method to capture semantic features for open-vocabulary 3D scene understanding.}}
    \label{fig:scannet}
\end{figure*}

\begin{table*}[htbp]
\centering
\caption{\textnormal{Performance of semantic segmentation on the Scannet dataset compared to SOTA methods based on text query.}}
\label{tab:scannet}
\begin{tabular}{l|cccccc}
\toprule
\multirow{2}{*}{Methods} & \multicolumn{2}{c}{19 classes} & \multicolumn{2}{c}{15 classes} & \multicolumn{2}{c}{10 classes} \\
& mIoU $\uparrow$ & mAcc. $\uparrow$ & mIoU $\uparrow$ & mAcc. $\uparrow$ & mIoU $\uparrow$ & mAcc. $\uparrow$ \\
\midrule
LangSplat~\cite{qin2024langsplat} & 2.0 & 9.2 & 4.9 & 14.6 & 8.0 & 23.9 \\
LEGaussians~\cite{shi2024language} & 1.6 & 7.9 & 4.6 & 16.1 & 7.7 & 24.9 \\
OpenGaussian~\cite{wu2024opengaussian} & \underline{30.1} & \underline{46.5} & 38.1 & 56.8 & \underline{49.7} & \underline{71.4} \\
Dr.Splat~\cite{jun2025dr} & 28.0 & 44.6 & \underline{38.2} & \underline{60.4} & 47.2 & 68.9 \\
CAGS(Ours) & \textbf{32.6} & \textbf{48.9} & \textbf{41.1} & \textbf{62.0} & \textbf{54.8} & \textbf{75.9} \\
\bottomrule
\end{tabular}
\end{table*}

\subsection{Open-Vocabulary Object Selection in 3D Space}

\textbf{Settings.} 1) \textit{Task}: For a given open-vocabulary text query, we utilize CLIP to extract its text feature and compute the cosine similarity with the language features of each Gaussian. Subsequently, we identify the most relevant 3D points and render them into multi-view images through the 3DGS pipeline. 2) \textit{Baseline}: Our method is evaluated against LangSplat, LEGaussians, OpenGaussian, and Dr.Splat. In the cases of LangSplat and LEGaussians, we adhered to their procedure to reconstruct the 512-dimensional CLIP feature from each Gaussian’s low-dimensional language feature. We note that all our evaluations maintain a uniform setup: text is used to identify matching 3D Gaussians, which are then rendered into multi-view images. The reported metrics align with the official metrics of the comparison methods. 3) \textit{Dataset and Metrics}: Experiments are performed on the LERF-OVS dataset, re-annotated by LangSplat. We calculate the average IoU and accuracy by comparing the images rendered from the 3D Gaussian points, selected based on the text query, with the ground truth object masks.

\noindent\textbf{Results.} We compare CAGS quantitatively with 3DGS-based language-embedded models on the LERF-ovs dataset, as shown in Tab.~\ref{tab:lerf}. CAGS outperforms the SOTA, Dr.Splat, by over 7 in mIoU and 5 in mAcc on average, with notable gains in complex scenes like \textit{teatime} and \textit{figurines}. OpenGaussian often exhibits false activations, highlighting its limited feature discrimination. In contrast, CAGS leverages context-aware modeling to ensure feature distinctiveness, avoiding the 2D-3D inconsistencies seen in OpenGaussian and Dr. Splat. Qualitative results in Fig.~\ref{fig:lerf} show that CAGS renders objects with clearer boundaries, unlike OpenGaussian, which struggles with ambiguous feature representations.

\subsection{Open-Vocabulary Point Cloud Understanding}

\textbf{Settings.} 1) \textit{Task}: For a set of open-vocabulary text queries, we compute the cosine similarity between the text features and the Gaussian features, assigning each Gaussian the category with the highest similarity to form the open-vocabulary point cloud task. 2) \textit{Baseline}: The comparison methods include LangSplat, LEGaussians, OpenGaussian, and Dr.Splat. 3) \textit{Dataset and Metrics}: We perform comparisons on the ScanNetv2 dataset~\cite{dai2017scannet}, which includes posed RGB images from video scans, reconstructed point clouds, and ground truth 3D point-level semantic labels. Both our approach and the baselines initialize with the provided point clouds. During training, we fix the point cloud coordinates and deactivate the 3DGS densification process to ensure the output point clouds’ number and coordinates align with the input and ground truth point clouds. For evaluation, we select the same 10 scenes as those chosen in Dr.Splat, extracting training images every 20 frames from the provided video images. We adopt point cloud mIoU and mAcc as the evaluation metrics.

\noindent\textbf{Results.} Quantitative results on the ScanNetv2 dataset are shown in Tab.~\ref{tab:scannet}, demonstrating that CAGS significantly outperforms SOTA methods across 19, 15, and 10-class settings. OpenGaussian and Dr. Splat struggle to distinguish sparse point clouds (e.g., mislabeling nearby objects), due to their per-Gaussian feature limitations. In contrast, CAGS’s context-aware propagation ensures robust feature aggregation, enabling precise segmentation. Point cloud feature visualizations on the ScanNet dataset, presented in Fig.~\ref{fig:scannet}, demonstrate that our CAGS method achieves enhanced instance-level separation.

\begin{table}[tbp]
    \centering
    \caption{\textnormal{Ablation study on the lerf dataset}}
    \label{tab:ablation}
    \begin{tabular}{l|cc}
        \toprule
        \multirow{2}{*}{Methods} & \multicolumn{2}{c}{Metrics} \\
        & mIoU $\uparrow$ & mAcc. $\uparrow$ \\
        \midrule
        Per-Gaussian baseline \cite{wu2024opengaussian} & 38.36 & 51.43 \\
        Our baseline solution  (HDBSCAN) & 37.02 & 49.21 \\
        + mask-aware contrastive learning & 40.32 & 54.31 \\
        + neighborhood feature aggregation & 47.97 & 66.86 \\
        + global feature propagation (full method) & \textbf{50.79} & \textbf{69.62} \\
        \bottomrule
    \end{tabular}
\end{table}

\subsection{Ablation Study}
We perform an ablation study to assess the impact of each component in our proposed CAGS framework on the LERF-OVS dataset. The quantitative results, shown in Table~\ref{tab:ablation}, demonstrate that our baseline solution using HDBSCAN clustering slightly underperforms the per-Gaussian baseline, but performance improves significantly with the addition of each proposed component, underscoring the effectiveness of our approach.

\noindent\textbf{Contextual Feature Propagation Network}
We ablate the number of layers in the Contextual Feature Propagation Network on the LERF-OVS dataset, as shown in Table~\ref{tab:layer}. Starting with direct feature training (0 layers), performance improves with the addition of one layer and peaks at two layers, but declines with three and four layers, while memory usage increases progressively. This trend occurs because contextual propagation aggregates spatial information to reduce cross-view granularity inconsistency, but excessive layers over-smooth features, diluting fine-grained details critical for 3D scene understanding, thus confirming two layers as the optimal choice for balancing effectiveness and efficiency in CAGS.

\noindent\textbf{Contrastive Learning Strategy}
In our ablation study of the Mask-Aware Contrastive Learning module on the LERF-OVS dataset, as shown in Table~\ref{tab:loss}, we added cohesion to the per-pixel InfoNCE loss for a fair comparison, where cohesion loss is a mechanism introduced by OpenGaussian that enforces the consistency of each point's feature within a mask with the mean feature of that mask. We found that incorporating cohesion loss from OpenGaussian does enhance the performance of per-pixel InfoNCE. Surprisingly, when we added cohesion loss to per-mask InfoNCE, performance decreased. Here's the rationale: per-mask InfoNCE emphasizes instance-level separation, relying on centroid-based learning to reduce noise from inconsistent segmentations. Adding cohesion over-constrains intra-mask features. This over-constraint upsets the balance that centroid-based learning has against the inconsistent fragmentations from SAM, weakening the robustness of this learning method and thus degrading performance.

\begin{table}[tbp]
\centering
\caption{\textnormal{Loss ablation}}
\label{tab:loss}
\begin{tabular}{c|c|cc}
\toprule
Infonce Loss & Cohesion Loss & mIoU $\uparrow$ & mAcc. $\uparrow$ \\
\midrule
Per-Pixel & \checkmark  & 44.24 & 65.43 \\
Per-Pixel &   & 43.34 & 63.82 \\
Feature-Mean(Per-Mask)  & \checkmark & 47.12 & 66.93 \\
Feature-Mean(Per-Mask)  &  & \textbf{50.79} & \textbf{69.62} \\
\bottomrule
\end{tabular}
\end{table}\textbf{
}

\begin{table}[tbp]
    \centering
    \caption{\textnormal{Ablation Study on the Number of Layers in Contextual Feature Propagation Network. (0 layer means direct feature training without Propagation Network)}}
    \label{tab:layer}
    \begin{tabular}{c|cc|c}
        \toprule
        \multirow{2}{*}{Propagation Layers} & \multicolumn{2}{c|}{Metrics} & \multirow{2}{*}{Memory(GB)} \\
        & mIoU $\uparrow$ & mAcc. $\uparrow$ &  \\
        \midrule
        0 Layer & 40.32 & 54.31 & 0 \\
        1 Layers & 47.92 & 66.34 & 1.4 \\
        2 Layers & \textbf{50.79} & \textbf{69.62} & 2.1 \\
        3 Layers & 49.24 & 68.72 & 2.8 \\
        4 Layers & 46.81 & 65.14 & 3.6 \\
        \bottomrule
    \end{tabular}
\end{table}

%% file: sec/6_conclusion.tex
\section{Conclusion}
We addressed cross-view granularity inconsistency in open-vocabulary 3D scene understanding using 3D Gaussian Splatting (3DGS), a challenge that fragments object representations across views and complicates accurate semantic interpretation. Our \textit{Context-Aware Gaussian Splatting (CAGS)} framework effectively mitigates this issue through three key contributions: a contextual feature propagation mechanism to reduce noise from inconsistent granularity by leveraging spatial relationships, a mask-centric contrastive learning strategy to smooth SAM-derived features across views and ensure semantic consistency, and a precomputation strategy that reduces computational cost by precomputing neighborhood relationships for efficient training. CAGS significantly enhances 3D instance segmentation by producing more cohesive feature representations, as evidenced by reduced fragmentation errors on challenging datasets like LERF-OVS and ScanNet. These improvements demonstrate the critical value of spatial context in achieving robust 3D understanding, particularly in complex multi-view scenarios. This work enables more robust language-guided interactions in 3D environments, facilitating seamless integration with language models for tasks like scene querying and object identification. Such advancements have broad applications in fields like robotics, where precise 3D understanding supports autonomous navigation, and augmented reality, where consistent scene representations enhance user immersion.